\documentclass{article} 
\usepackage{nips15submit_e,times}
\usepackage{hyperref}
\usepackage{graphicx}
\usepackage{url}
\usepackage{amsmath}
\usepackage{multirow}
\usepackage{amssymb}
\usepackage{amsmath}
\usepackage{caption}
\usepackage{subcaption}
\usepackage{epstopdf}
\nipsfinalcopy

\title{A Deep Hashing Learning Network}

\author{Guoqiang zhong, Pan Yang, Sijiang Wang, Junyu Dong \\
Ocean University of China\\
Qing Dao, China \\
\texttt{young.ouc@gmail.com}}

\begin{document}

\maketitle

\begin{abstract}
Hashing-based methods seek compact and efficient binary codes that preserve the neighborhood structure in the original data space. For most existing hashing methods, an image is first encoded as a vector of hand-crafted visual feature, followed by a hash projection and quantization step to get the compact binary vector. Most of the hand-crafted features just encode the low-level information of the input, the feature may not preserve the semantic similarities of images pairs. Meanwhile, the hashing function learning process is independent with the feature representation, so the feature may not be optimal for the hashing projection. In this paper, we propose a supervised hashing method based on a well designed deep convolutional neural network, which tries to learn hashing code and compact representations of data simultaneously. The proposed model learn the binary codes by adding a compact sigmoid layer before the loss layer. Experiments on several image data sets show that the proposed model outperforms other state-of-the-art methods.
\end{abstract}

\section{Introduction}

Similarity search generally involves a large scale collection of data (e.g. images, videos, documents) that are represented as points in a high dimensional space. We are required to find the most similar (top-k nearest) instance to the query. This is the most important role for the search engine, as well as the areas such as data compression and pattern recognition. It has various applications in real world, for example: scene completion [1], image retrieval, plagiarism analysis [2] and so on.

For most existing hashing methods, an input is first projected into a low-dimensional subspace, then followed by a quantization step to get the compact binary vector. Locality Sensitive Hashing (LSH) and its extensions [3,4,5,6] based on randomized projections are one of the most widely employed hashing methods in industrial practice solving ANN (approximate nearest neighbor) search. The most magnitude advantage of this technique is that the random projects can maintain the similarity of pairs in original data space provably, meanwhile, the random initialization of projection matrix do not need extra computation, This makes LSH suitable for large scale ANN tasks. However, higher precision in general require long codes which lead to low recall and more storage cost.

In contrast to the data-independent hash framework employed in LSH-related methods, most of recent research focuses on data-dependent hashing which learns projection function from training data. Semantic hashing [7] uses a deep graphical model to learn the hash function, by forcing a deep layer to be small. Anchor graph hashing [8] and spectral hashing [9] use spectral graph partitioning for hashing with the graph constructed from data similarity relationship. Multidimensional spectral hashing [10] introduces a new formulation which seeks to reconstruct the affinity between datapoints, rather than the distance.  Binary reconstruction embedding [11] learns hash function by explicitly minimizing the reconstruction error between the original distances and the Hamming distances of the corresponding binary embeddings. Minimal loss hashing [12] formulates the hashing problem as a structured prediction problem with latent variables and a hinge-like loss function. PCA-ITQ (Iterative quantization) [13,14] one recent data-dependent method which outperforms most other state-of-the-art approaches, this method finds an orthogonal rotation matrix to refine the initial projection matrix learned by principal component analysis (PCA), so that the quantization error of mapping data to the vertices of binary hypercube is minimized.

All of the hashing methods we mentioned above learn hash function based on some hand-crafted visual descriptors (e.g. GIST [15], BoF [16,17]). However, those hand-crafted features can just extract the low-level representation of the data, which may not be able to preserve the semantic similarities of image pairs. At the same time, the hash function learning procedure is independent with the feature extraction step. Hence, hash code learning can not give feedback to the feature representation learning step, and vice versa.

In this paper, we introduce a supervised hashing method based on a well designed deep convolutional neural network, which combines the feature learning and hashing function learning together. We have compared our model with multiple hashing approaches. The results show that our method can achieve state-of-the-art, even better performance for image retrieval.

\section{Related Work}
\label{gen_inst}

Convolutional neural networks (CNNs) [18,19] have demonstrated its great power in several visual recognition field, and exceeds human-level performance in many tasks, including recognizing traffic signs [20], faces [21,22], and hand written digits [20,23]. Meanwhile, the deep convolutional neural network based approaches have recently been substantially improved on state-of-the-art in large scale image classification [24,25], object detection [26,27], and many other tasks [28,29].

Compared with the conventional hand-craft visual descriptors, which are designed by human engineers with an unsupervised fashion for specific task. Deep convolutional neural networks encode images into multiple levels of representation. With those suitable representation, we can discover complex structures hidden behind high dimensional data. The key to the success of such deep architecture is their ability of representation learning. For classification tasks, the higher layers of representation reserve the important aspects of the input for discrimination and inhibit the irrelevant variations. In this work, based on their great representation learning capability, we utilize the deep CNNs to automatically learn image feature instead of using hand-craft feature(e.g.,Gist, Bof).

Currently, as the great success made by machine learning on many tasks, numerous models have been introduced to hashing applications. Semantic hashing [31] introduce a probabilistic model to learn the marginal distribution over input vector. The assumptions used in semantic hashing fit the constraint in equation \ref{equ1} ideally. However, semantic hashing needs complex and difficult to train the network. CNN hashing [32] is a two stage hash method to learn optimal hashing code. In the first stage, an approximate hashing code is learned by decomposing the similarity matrix $S$ into a product form $S \approx \frac{1}{q}HH^T$. The k-th row in $H$ is regarded as the approximate target hashing code, then the learned hashing code is assigned as the supervised information to learn the hash function. This two stage framework leads to good performance. However, the matrix decomposition limits the scale of the application. Meanwhile, the learned image representation can not give feedback for learning better approximate hash code. We propose a method that can combine feature learning and hashing code learning together. This end-to-end architecture improves previous state-of-the-art supervised and unsupervised hashing methods.

\section{The Proposed Model}
\label{headings}

\subsection{Hash Statement}

Generally speaking, a good code for hashing satisfies three conditions [30]:(1) projecting similar pairs in data space to similar binary codewords in Hamming space (2) a small number of bits to encode each sample (3) little computation for input projection. For a given data set $\{x_1,x_2,...,x_n\}$ with $x_i \in R^d$, let $\{y_i\}_{i=1}^n$ with $y_i \in \{0,1\}^m$ be the binary code for each input. In general, we assume different bits are independent of each other, that is $y_i = [h_1 (x_i), h_2 (x_i), ..., h_m (x_i) ]^T$ with $m$ independent binary hashing functions $\{h_k (.)\}_{k=1}^m$. We require that each bit has a 50\% chance of being one or zero. Our goal is to minimize the average Hamming distance between similar pairs, we obtain the following problem with the goal:
\begin{equation} \label{equ1}
\begin{split}
minimize: \sum_{ij} W_{ij} || y_i - y_j ||^2 \\
s.t. \ y_i \in \{0,1\}^k \\
\sum_i y_i  = 0 \\
\frac{1}{n} \sum_i y_i y_i^T  = I
\end{split}
\end{equation}
where the constraint $\sum_i y_i = 0$ corresponds to the 50\% probability assumption, while the constraint $\frac{1}{n}\sum_i y_i y_i^T$ corresponds to the independent assumption of bits.

\begin{figure}[h]
    \centering
    \includegraphics[width=0.8\textwidth]{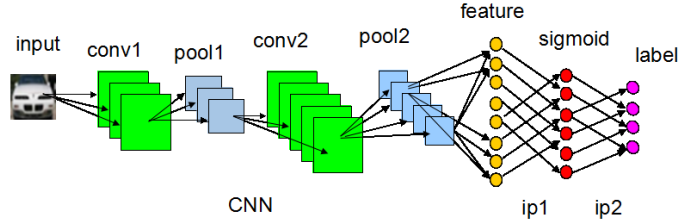}
    \caption{Illustration of the end-to-end deep hashing learning network.}
    \label{fig:img_net}
\end{figure}

In the following section, we describe the model in detail. Figure \ref{fig:img_net} shows a example of the pipline of the deep convolutional neural network, a linear hash projection layer is followed by a sigmoid quantization layer, the network was trained use backpropagation.

Denotes $\mathcal{X}$ to be the image space,$\mathcal{X} = \{x_1,x_2, ..., x_n\}$, our goal of hash learning for images is to find a projection $\mathcal{H}:\mathcal{X}  \rightarrow \{0,1\}^m$. because it is NP hard to compute the best binary functions $ h_k(.) $ for image set $ \mathcal{X} $ [9], hashing methods adopt a two-stage strategy to learn $h_k(.)$, the project stage by $m$ real-value hash functions $ \{h_k(x)\}_{k=1}^m $ and the quantization stage by threshold the real-values to binary.

\subsection{Model Formulation}
In standard locality sensitive hashing, each hash function $h_k$ is generated independently by setting a random vector $ l_k $ from a gaussian distribution with zero mean and identity covariance. Then the hash function can be expressed as $h_p(x) = sign(l_p^T x)$. In our approach, the input image was first been mapping to the feature vector, with multiple convolution and pooling operation,
\begin{equation} \label{equ2}
    \begin{split}
        h_k(x) = sigmoid \left(\sum_{j=1}^m W_{j.} (CNN(x)) \right)
    \end{split}
\end{equation}
where $m$ denotes the number of hash function, $CNN(x)$ denotes the feature extraction on the input images, $W_{j.}$ is the projection vector for the $k-th$ hash function. Each hash function $h_k(.)$ is learned independently by put a linear mapping on the same feature representation layer.

Sigmoid function refers to the special case of the logistic function, which has an "S" shape, due to its easily calculated derivatives and physiological characteristics, sigmoid function was widely used in neural networks until the ReLU(rectified linear units) and its extensions get widely used. Sigmoid function can compress the signal to $[0,1]$, experiments in our work show that the output of sigmoid layer most distribute at the tailer of the sigmoid curve, which has near zero or near one value.

CNN achieve great success on image classification tasks, the major drawback of many feature learning structure is their complexity, those alforithms usually require a careful selection of multiple hyperparameters, such as learning rates, momentum, weight decay, a good initialization of filter also be a key to good performance. In this paper, we adopts a simple framework of CNN for the needs of fast nearest neighbor search.

\section{Experiments}

\subsection{Data Sets}
To evaluate the proposed method, we use two benchmark datasets with different kinds of images, MNIST [33]\footnote{http://yann. lecun. com/exdb/mnist}, and CIFAR-10 [35]\footnote{http://www.cs.toronto.edu/~kriz/cifar.html}. The first dataset is MNIST, with 70K 28*28 greyscale images of hand written digits from zero to nine, has a training set of 60,000 examples, and a test set of 10,000 examples.The second dataset is CIFAR-10, consists of 60000 32x32 colour images in 10 classes of natural objects, with 6000 images per class. There are 50000 training images and 10000 test images.

\subsection{Baselines}

In this paper, four most representative unsupervised methods, PCA-ITQ, LSH, SH, PCAH and 2 supervised methods KSH, BRE are chosen to compare the performance of the proposed hash methods.

We randomly chose 1000 images as the query sequence, For unsupervised methods, all of the rest images as training samples, for supervised methods, we take original training set as training samples. For the proposed method, we directly use the image pixels as input, for the baseline methods, we follow the general setting to get the image representations, use 784-dimensional grayscale vector for MNIST image, and 512-dimensional GIST vector for CIFAR-10. We mean-centered the data and normalized the feature vector to have unit norm. We adopt the scheme widely used in image retrieval task to evaluate our methods, including mean average precision, precision-recall curves, precision curves within hammming distance and precision curves w.r.t number of top returned images.

\subsection{Model Configuration}

We implements the proposed methods based on open source Caffe [37] framework. we use 32, 32, 64 filters with size 5*5 in the first, second, and third convolutional layers, with each followed by the ReLU activation function. The hash mapping layer located at the top of the third pooling layer, then a compression sigmoid layer is followed.

\begin{table}[h]
\caption{mAP on MNIST and CIFAR-10 dataset, w.r.t different number os bits}
\label{my-label}
\resizebox{\textwidth}{!}{%
\begin{tabular}{|c|l|l|l|l|l|l|l|l|}
\hline
\multirow{2}{*}{method} & \multicolumn{4}{c|}{MNIST(MAP)}   & \multicolumn{4}{c|}{CIFAR-10(MAP)} \\ \cline{2-9}
                        & 16bits & 24bits & 32bits & 48bits & 16bits  & 24bits & 32bits & 48bits \\ \hline
LSH                     &   0.250    &    0.284     &    0.310    &    0.430    &    0.298     &     0.344   &   0.331    &    0.389    \\ \hline
SH                      &   0.347    &    0.383     &    0.393    &    0.387    &    0.352     &     0.355   &   0.379    &    0.381    \\ \hline
PCAH                    &   0.351    &    0.344     &    0.332    &    0.309    &    0.291     &     0.280   &   0.272    &    0.261    \\ \hline
PCA-ITQ                 &   0.515    &    0.550     &    0.581    &    0.610    &    0.427     &     0.445   &   0.453    &    0.469    \\ \hline
SKLSH                   &   0.182    &    0.231     &    0.218    &    0.256    &    0.288     &     0.312   &   0.334    &    0.394    \\ \hline
KSH                     &   -    &    0.891     &    0.897    &    0.900    &    0.303     &     0.337   &   0.346    &    0.356    \\ \hline
BRE                     &   -    &    0.593     &    0.613    &    0.634    &    0.159     &     0.181   &   0.193    &    0.196    \\ \hline
CNN+                    &   -    &    0.975     &    0.971    &    0.975    &    0.465     &     0.521   &   0.521    &    0.532    \\ \hline
Ours                    &   \textbf{0.992}     &   \textbf{0.993}     &    \textbf{0.995}     &    \textbf{0.993}    &    \textbf{0.714}     &  \textbf{0.718}      &   \textbf{0.736}     &    \textbf{0.728}    \\ \hline
\end{tabular}
}
\end{table}

\subsection{Results Analysis}

\begin{figure}[h]

\begin{subfigure}{0.5\textwidth}
\includegraphics[width=0.9\linewidth, height=5cm]{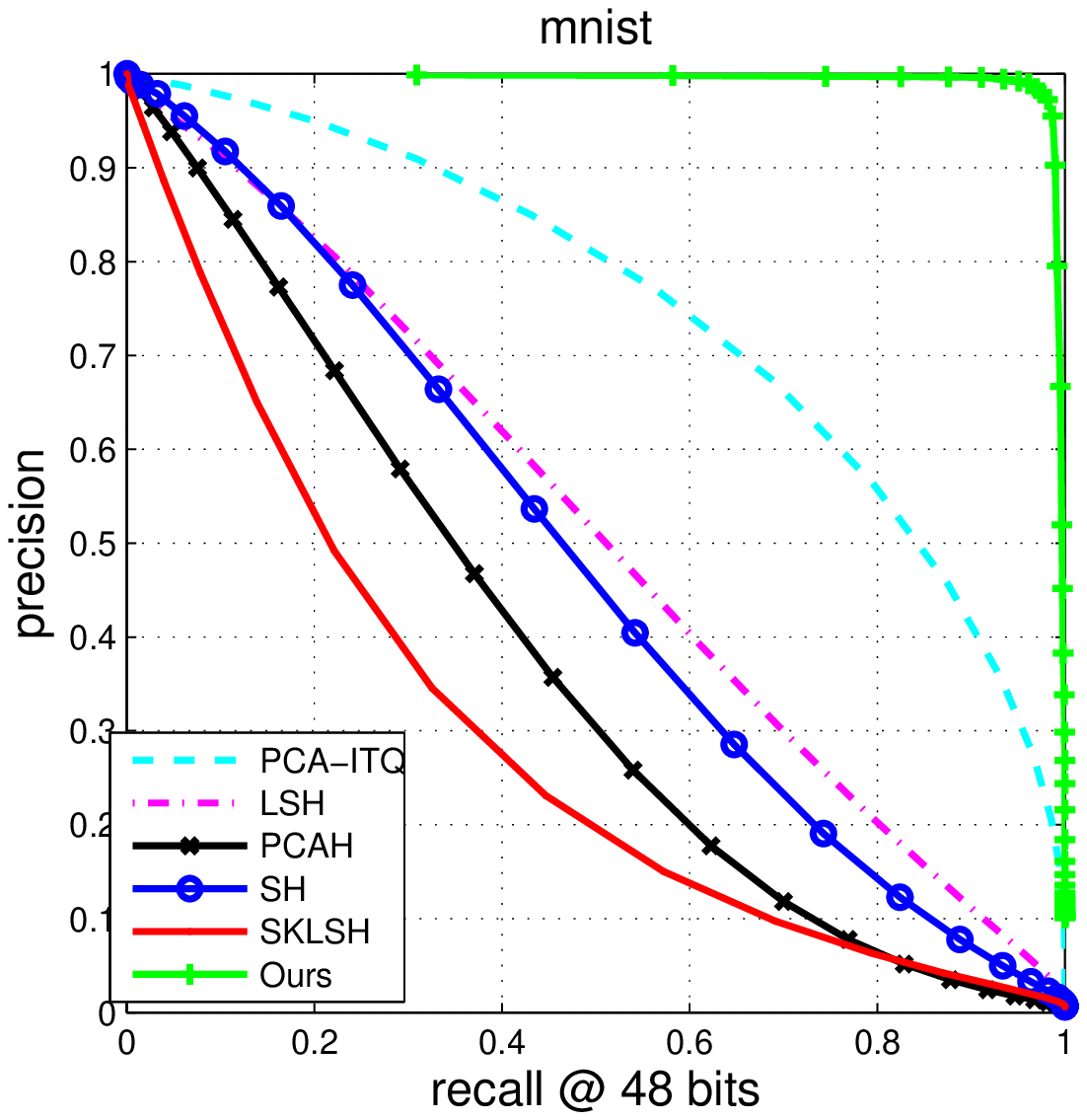}
\caption{}
\label{fig:subim1}
\end{subfigure}
\begin{subfigure}{0.5\textwidth}
\includegraphics[width=0.9\linewidth, height=5cm]{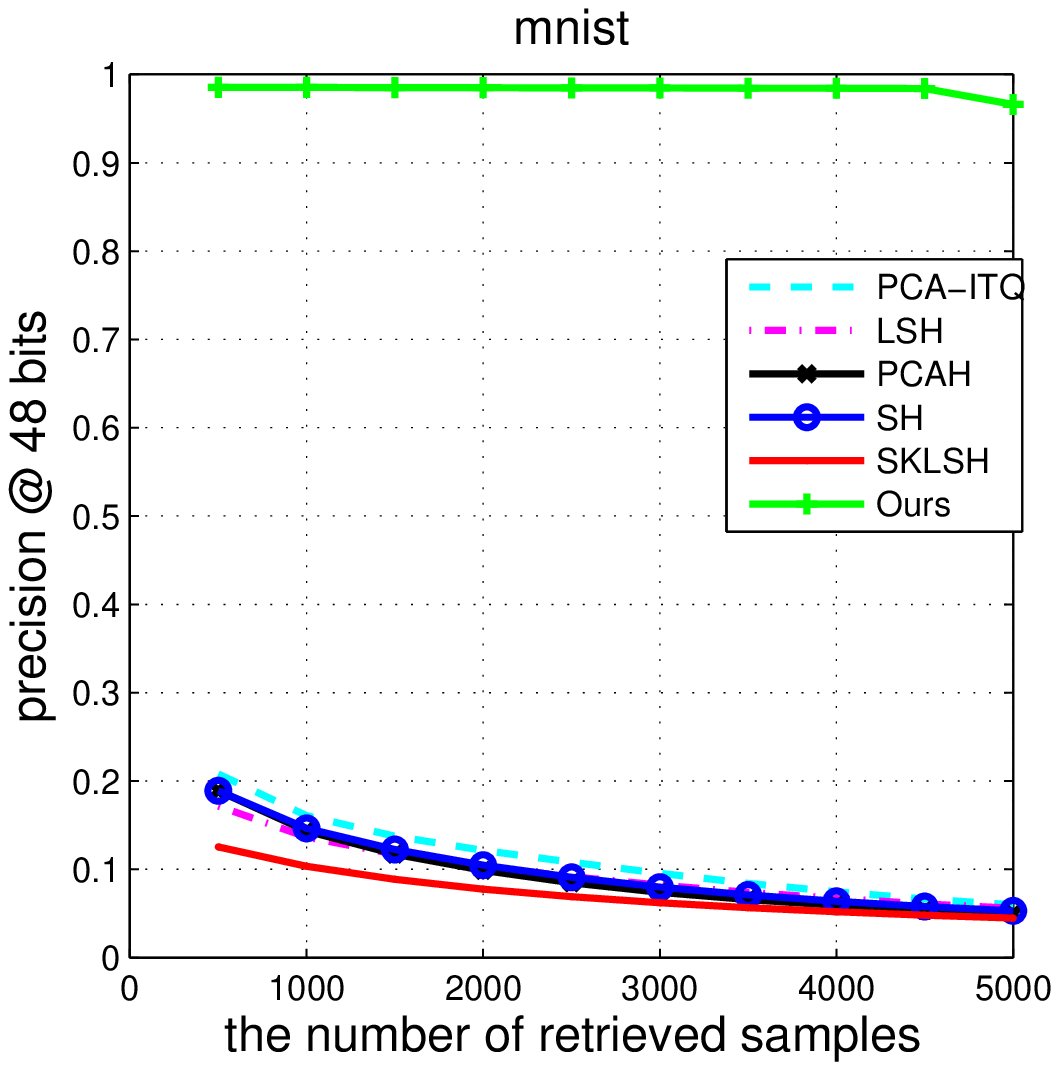}
\caption{}
\label{fig:subim2}
\end{subfigure}

\caption{Quantitative comparison results on CIFAR-10. (a) Precision-recall curves with 48 bits, (b) Precision curves w.r.t numbers of top returned images}
\label{result_mnist}
\end{figure}

\begin{figure}[h]

\begin{subfigure}{0.5\textwidth}
\includegraphics[width=0.9\linewidth, height=5cm]{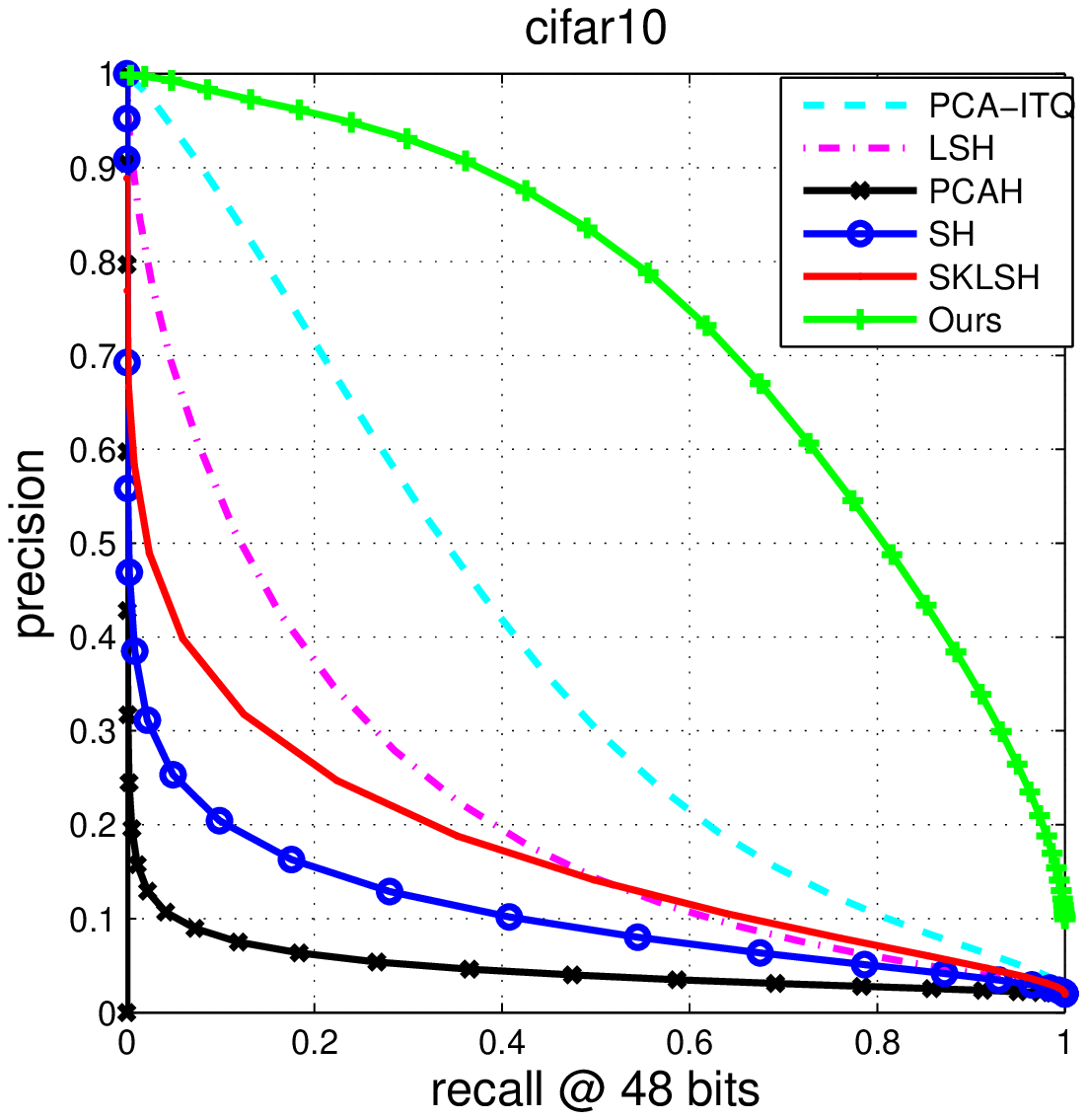}
\caption{}
\label{fig:subim3}
\end{subfigure}
\begin{subfigure}{0.5\textwidth}
\includegraphics[width=0.9\linewidth, height=5cm]{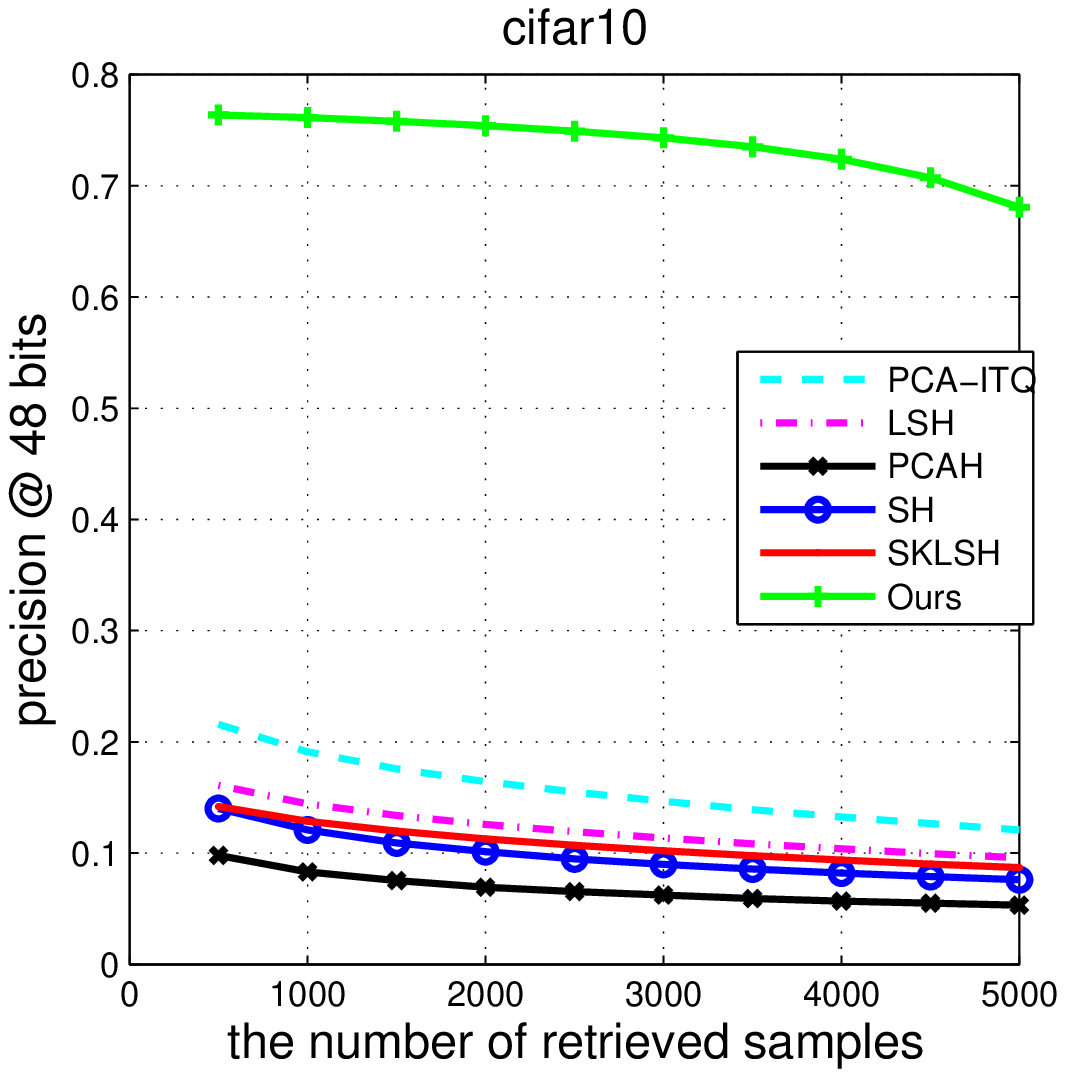}
\caption{}
\label{fig:subim4}
\end{subfigure}

\caption{Quantitative comparison results on CIFAR-10. (a) Precision-recall curves with 48 bits, (b) Precision curves w.r.t numbers of top returned images}
\label{result_cifar}
\end{figure}

Table \ref{my-label} and Figure \ref{result_mnist} to \ref{result_cifar} show the precision-recall curves and other two evaluation curves comparison on the evaluate datasets, all of the unsupervised methods are obtained by the open source implementations provided by their respective authors, we directly use the results of the supervised methods KSH and BRE obtained by [32], KSH need extra time for k-means learning, with respect to large scale data, the hashing learning may suffer the problem of time consuming. We also compare the mAP result with CNNH, the MAP result of our method gains 0.2\% w.r.t to CNNH on MNIST. Particularly, our model indicate a increase of 18\% - 27\% on CIFAR-10 w.r.t state-of-the-art method, The substantial superior performance verifies the efficiency of our end-to-end framework.

Compared to the conventional methods, CNN based methods can achieve much better result, which we think is the influence of the automatically learned image representation. As we mentioned before, good hashing code satisfies the requirement of similarity preserve, less bits and little computation. Hence, any time consuming computing should be avoiding. In this work, we adopt a simple CNN to learn feature representation and hashing code, more complex model can promote the performance, but the cost for fast similarity search will increase as well.

\section{Conclusion}
In this paper, we proposed a end-to-end supervised method for image retrieval, which simultaneously learns a compact hash code as well as a good feature representation of images. This method has no restrict on data scale and can generate hash code with little computation, the model can be boosted by GPU acceleration and multithreading. The proposed method learn the hash code with the image label, we just use some simple CNN model to learn the hash code, experiments show that the retrieval results can be promoted by more powerful classification model. Even with such simple model, our method has astonishing performance gains over state-of-the-arts.


\subsubsection*{References}

\small{
[1] Hays, J. \& Efros, A. A. (2007). Scene completion using millions of photographs. ACM Transactions on Graphics (TOG), 26(3), 4.

[2] Stein, B., zu Eissen, S. M. \& Potthast, M. (2007, July). Strategies for retrieving plagiarized documents. In Proceedings of the 30th annual international ACM SIGIR conference on Research and development in information retrieval (pp. 825-826). ACM.

[3] Datar, M., Immorlica, N., Indyk, P. \& Mirrokni, V. S. (2004, June). Locality-sensitive hashing scheme based on p-stable distributions. In Proceedings of the twentieth annual symposium on Computational geometry (pp. 253-262). ACM.

[4] Kulis, B. \& Grauman, K. (2009, September). Kernelized locality-sensitive hashing for scalable image search. In Computer Vision, 2009 IEEE 12th International Conference on (pp. 2130-2137). IEEE.

[5] Kulis, B., Jain, P. \& Grauman, K. (2009). Fast similarity search for learned metrics. Pattern Analysis and Machine Intelligence, IEEE Transactions on, 31(12), 2143-2157.

[6] Mu, Y. \& Yan, S. (2010, March). Non-Metric Locality-Sensitive Hashing. In AAAI.

[7] Salakhutdinov, R. \& Hinton, G. (2009). Semantic hashing. International Journal of Approximate Reasoning, 50(7), 969-978.

[8] Liu, W., Wang, J., Kumar, S. \& Chang, S. F. (2011). Hashing with graphs. In Proceedings of the 28th International Conference on Machine Learning (ICML-11) (pp. 1-8).

[9] Weiss, Y., Torralba, A. \& Fergus, R. (2009). Spectral hashing. In Advances in neural information processing systems (pp. 1753-1760).

[10] Weiss, Y., Fergus, R. \& Torralba, A. (2012). Multidimensional spectral hashing. In Computer Vision¨CECCV 2012 (pp. 340-353). Springer Berlin Heidelberg.

[11] Kulis, B. \& Darrell, T. (2009). Learning to hash with binary reconstructive embeddings. In Advances in neural information processing systems (pp. 1042-1050).

[12] Norouzi, M. \& Blei, D. M. (2011). Minimal loss hashing for compact binary codes. In Proceedings of the 28th International Conference on Machine Learning (ICML-11) (pp. 353-360).

[13] Gong, Y. \& Lazebnik, S. (2011, June). Iterative quantization: A procrustean approach to learning binary codes. In Computer Vision and Pattern Recognition (CVPR), 2011 IEEE Conference on (pp. 817-824). IEEE.

[14] Gong, Y., Lazebnik, S., Gordo, A. \& Perronnin, F. (2013). Iterative quantization: A procrustean approach to learning binary codes for large-scale image retrieval. Pattern Analysis and Machine Intelligence, IEEE Transactions on, 35(12), 2916-2929.

[15] Oliva, A. \& Torralba, A. (2006). Building the gist of a scene: The role of global image features in recognition. Progress in brain research, 155, 23-36.

[16] Csurka, G., Dance, C., Fan, L., Willamowski, J. \& Bray, C. (2004, May). Visual categorization with bags of keypoints. In Workshop on statistical learning in computer vision, ECCV (Vol. 1, No. 1-22, pp. 1-2).

[17] Sivic, J. \& Zisserman, A. (2003, October). Video Google: A text retrieval approach to object matching in videos. In Computer Vision, 2003. Proceedings. Ninth IEEE International Conference on (pp. 1470-1477). IEEE.

[18] Krizhevsky, A., Sutskever, I. \& Hinton, G. E. (2012). Imagenet classification with deep convolutional neural networks. In Advances in neural information processing systems (pp. 1097-1105).

[19] LeCun, Y., Boser, B., Denker, J. S., Henderson, D., Howard, R. E., Hubbard, W. \& Jackel, L. D. (1989). Backpropagation applied to handwritten zip code recognition. Neural computation, 1(4), 541-551.

[20] Maas, A. L., Hannun, A. Y. \& Ng, A. Y. (2013). Rectifier nonlinearities improve neural network acoustic models. In Proc. ICML (Vol. 30).

[21] Sun, Y., Chen, Y., Wang, X. \& Tang, X. (2014). Deep learning face representation by joint identification-verification. In Advances in Neural Information Processing Systems (pp. 1988-1996).

[22] Taigman, Y., Yang, M., Ranzato, M. A. \& Wolf, L. (2014, June). Deepface: Closing the gap to human-level performance in face verification. In Computer Vision and Pattern Recognition (CVPR), 2014 IEEE Conference on (pp. 1701-1708). IEEE.

[23] Wan, L., Zeiler, M., Zhang, S., Cun, Y. L. \& Fergus, R. (2013). Regularization of neural networks using dropconnect. In Proceedings of the 30th International Conference on Machine Learning (ICML-13) (pp. 1058-1066).

[24] Sermanet, P., Eigen, D., Zhang, X., Mathieu, M., Fergus, R. \& LeCun, Y. (2013). Overfeat: Integrated recognition, localization and detection using convolutional networks. arXiv preprint arXiv:1312.6229.

[25] Chatfield, K., Simonyan, K., Vedaldi, A. \& Zisserman, A. (2014). Return of the devil in the details: Delving deep into convolutional nets. arXiv preprint arXiv:1405.3531.

[26] Girshick, R., Donahue, J., Darrell, T. \& Malik, J. (2014, June). Rich feature hierarchies for accurate object detection and semantic segmentation. In Computer Vision and Pattern Recognition (CVPR), 2014 IEEE Conference on (pp. 580-587). IEEE.

[27] Zou, W. Y., Wang, X., Sun, M. \& Lin, Y. (2014). Generic object detection with dense neural patterns and regionlets. arXiv preprint arXiv:1404.4316.

[28] Gong, Y., Wang, L., Guo, R. \& Lazebnik, S. (2014). Multi-scale orderless pooling of deep convolutional activation features. In Computer Vision¨CECCV 2014 (pp. 392-407). Springer International Publishing.

[29] Zhang, N., Paluri, M., Ranzato, M. A., Darrell, T. \& Bourdev, L. (2014, June). Panda: Pose aligned networks for deep attribute modeling. In Computer Vision and Pattern Recognition (CVPR), 2014 IEEE Conference on (pp. 1637-1644). IEEE.

[30] Weiss, Y., Torralba, A. \& Fergus, R. (2009). Spectral hashing. In Advances in neural information processing systems (pp. 1753-1760).

[31] Salakhutdinov, R. \& Hinton, G. (2009). Semantic hashing. International Journal of Approximate Reasoning, 50(7), 969-978.

[32] Xia, R., Pan, Y., Lai, H., Liu, C. \& Yan, S. (2014, June). Supervised Hashing for Image Retrieval via Image Representation Learning. In Twenty-Eighth AAAI Conference on Artificial Intelligence.

[33] LeCun, Y. \& Cortes, C. (1998). The MNIST database of handwritten digits.

[34] Netzer, Y., Wang, T., Coates, A., Bissacco, A., Wu, B. \& Ng, A. Y. (2011). Reading digits in natural images with unsupervised feature learning. In NIPS workshop on deep learning and unsupervised feature learning (Vol. 2011, No. 2, p. 5). Granada, Spain.

[35] Krizhevsky, A. \& Hinton, G. (2009). Learning multiple layers of features from tiny images. Computer Science Department, University of Toronto, Tech. Rep, 1(4), 7.

[36] Norouzi, M. \& Blei, D. M. (2011). Minimal loss hashing for compact binary codes. In Proceedings of the 28th International Conference on Machine Learning (ICML-11) (pp. 353-360).

[37] Jia, Y., Shelhamer, E., Donahue, J., Karayev, S., Long, J., Girshick, R., ... \& Darrell, T. (2014, November). Caffe: Convolutional architecture for fast feature embedding. In Proceedings of the ACM International Conference on Multimedia (pp. 675-678). ACM.

\end{document}